\documentclass[conference]{IEEEtran}
\IEEEoverridecommandlockouts
\usepackage{cite}
\usepackage{amsmath,amssymb,amsfonts}
\usepackage[T1]{fontenc}
\usepackage{algorithmic}
\usepackage{listings}
\usepackage{graphicx}
\usepackage{textcomp}
\usepackage{xcolor}
\def\BibTeX{{\rm B\kern-.05em{\sc i\kern-.025em b}\kern-.08em
    T\kern-.1667em\lower.7ex\hbox{E}\kern-.125emX}}
\lstdefinestyle{sharpc}{language=[Sharp]C, frame=lr, rulecolor=\color{blue!80!black}}

\begin{document}

\title{Deep Learning and Artificial General Intelligence:
\\ Still a Long Way to Go}

\author{\IEEEauthorblockN{Maciej \'{S}wiechowski}
\IEEEauthorblockA{\textit{QED Software}, Warsaw, Poland \\
ORCID: 0000-0002-8941-3199 \\
maciej.swiechowski@qed.pl}}

\maketitle

\begin{abstract}
In recent years, deep learning using neural network architecture, i.e. deep neural networks, has been on the frontier of computer science research. It has even lead to superhuman performance in some problems, e.g., in computer vision, games and biology, and as a result the term deep learning revolution was coined. 

The undisputed success and rapid growth of deep learning suggests that, in future, it might become an enabler for Artificial General Intelligence (AGI). In this article, we approach this statement critically showing five major reasons of why deep neural networks, as of the current state, are not ready to be the technique of choice for reaching AGI. 
\end{abstract}

\begin{IEEEkeywords}
Deep Learning, Deep Neural Networks, Artificial Intelligence, Artificial General Intelligence, Machine Learning
\end{IEEEkeywords}

\section{Introduction}

The current applications of Artificial Intelligence (AI) belong to the so-called ``\emph{narrow AI}’’\cite{1}. \emph{Narrow AI} is often very successful but only in solving a particular task. For example, a top quality chess playing program \cite{2} is unable to form any sentence in a natural language.

Artificial General Intelligence (AGI) \cite{1, 3}, however, is the idea of creating multi-purpose AI that would be capable of learning and performing any intellectual tasks humans do. On a sidenote, a hypothetical AI that would surpass humans in a given domain is referred to as ``\emph{strong AI}’’. If such an AI would surpass humans in solving all problems that require intelligence, it would be called ``\emph{strong AGI}’’ and its existence would mean that the human development has achieved singularity.

AGI can be viewed as human-like intelligence displayed by machines. In fact, the most popular validation tests for AGI involve comparisons to humans in efficacy in solving tasks:
\begin{enumerate}
    \item \emph{Turing Test} \cite{4} – ability to carry out a believable conversation in a natural language. This test was proposed initially by Alan Turing in a different form and by the name of \emph{Imitation Game}. It is associated with one of the first research works on AI and computers. We also recommend a recent paper~\cite{swiechowski2020game} that discusses the importance of \emph{Imitation Game AI Competitions}.
    \item \emph{Robot College Student Test} \cite{3} – ability to enroll in a university, pass all exams and obtain a degree.
    \item \emph{Employment Test} \cite{5} – ability have a job which is ordinarily performed by humans and work as effectively in it as humans do.
    \item \emph{The Coffee Test} - ability to enter an ordinary American house (without any predefined setup) and make a cup of coffee. This includes figuring out where the cup as well as the coffee may be, mix all ingredients (sugar, water), etc. According to~\cite{goertzel2012architecture}, it was proposed by Steve Wozniak. 
\end{enumerate}

AGI may seem as something unreachable and vague. While it has not been achieved yet, it might be worth thinking of it as a distant research challenge. It has been one of the original long-term motivations driving the AI field. 

In this article, we are not to claim whether AGI will be possible or not. We focus on deep learning, which in the last decade, has become one of the major topics in research \cite{6}. Moreover, there are many spectacular achievements and commercial applications \cite{7} of deep learning. It is safe to say, that deep learning and deep neural networks (DNNs) have revolutionized the field \cite{8}.

This has raised the question – are DNNs the technological enabler that will lead us to AGI? 
In this article, we devote five sections to major reasons of why DNNs are not yet ready to be the technological driver for AGI. Due to strict page limits, we assume readers’ familiarity with the concepts of deep learning and neural networks, but not necessarily with AGI.

In order for this article to be technical and not philosophical, we omit such aspects of human thinking as sentience, self-awareness and motivation. We also do not theorize that having a body that can feel stimuli is a prerequisite for intelligence.

\section{Requirement for a lot of training data}

The process of training deep neural networks requires huge volumes of data \cite{9}. The amount of data necessary to reach high quality of a DNN model (e.g., expected accuracy of a classifier) depends on the complexity of a given problem and the size of the neural network used. The GPT-3 deep neural network, which is currently one of the biggest networks ever trained, has 175 billion learning parameters \cite{10}.

When a model has that many parameters that need to be optimized solely based on feeding them with training data, then we need to make sure that they are many training samples.

\noindent \textbf{We argue that this requirement for a large corpus of training data conflicts with the nature of AGI.}

Firstly, it is not feasible to have so much training data for any tasks.  AGI is inspired by human intelligence and most methods of testing whether we achieve it involve comparisons to humans. Humans can learn to be effective at virtually unlimited number of tasks. However, children do not need to observe a lot of various cats in order to be able to recognize a cat. While the learning by children has not been fully understood yet, it is most probable that a human’s brain builds an abstract internal representation of a concept relatively quickly \cite{11}. 

Secondly, large neural networks combined with large volumes of data result in high computational cost and, therefore, long training time \cite{12}. It would be completely ineffective to train for a long time for every task AGI is presented with. We need faster ways of training/creating AGI models.

\section{Transfer learning, using analogies and intuition }

Although transfer learning, using analogies and intuition mean different things, all of them may help in speeding up the learning process.

In essence, transfer learning \cite{13} is a paradigm in machine learning that consist in training a model for a given task and reusing it in a completely new task without or with minimal retraining. Multi-domain (or multi-task) learning is a step further and involves training one model for a variety of tasks. Finding similar concepts by analogy or reducing a problem to known ones by analogy is one way to implement transfer learning. Another way might be to provide an ontology that maps concepts from one task to another \cite{14}.

The problem with transfer learning is that there have not been any successful attempts on a very large scale i.e. for a lot of really diverse tasks. Readers are advised to refer to collective works and surveys on this topic, e.g. \cite{14,15}. We do not want to state that full multi-task learning is not possible with DNNs but rather that it is still in its early phase of research with very limited applications. 

Intuition is a concept attributed to human thinking, which has not been yet defined in terms of artificial intelligence. According to \cite{16}, intuitive solutions are found extremely quickly. They might not be optimal -- just first good approximations. Neural networks do not have such property. They work in one mode, i.e. the full potential, and they need to perform the necessary inference to give output signal. However, it may be argued that iterative algorithms such as Monte Carlo Tree Search \cite{17} or that can be asked at any time about the current solution to the problem are closer to displaying some aspects of intuition by means of early solutions.

\textbf{Humans display enormous capabilities to generalize experiences gained at one task into similar ones.} For example, when we learn how to cook a specific dish, we absorb a lot of abstract concepts about cooking, in general, that help us to prepare new dishes. Or we can read a manual of a game (e.g. a board game) that we never played before and be able to play it as long as it is based on concepts we have already encountered in other games. Furthermore, the ability to transfer knowledge and act in completely new situation is a strong trait of intelligence~\cite{piaget2003psychology}.

\section{Abstract out-of-the-box reasoning}

DNNs are universal function approximators. Ultimately, they map an input vector $\bar{x}$ into output vector:
\begin{equation}
f(\bar{x}): R^n \rightarrow  R^m
\end{equation}

The authors of \cite{18} say that machine learning is a modern rephrase of curve fitting. NNs learn on training examples and once they are trained they are able to generalize into new examples by similarity in how they activate the network.

\begin{figure}
\centering
\includegraphics[width=0.9\linewidth]{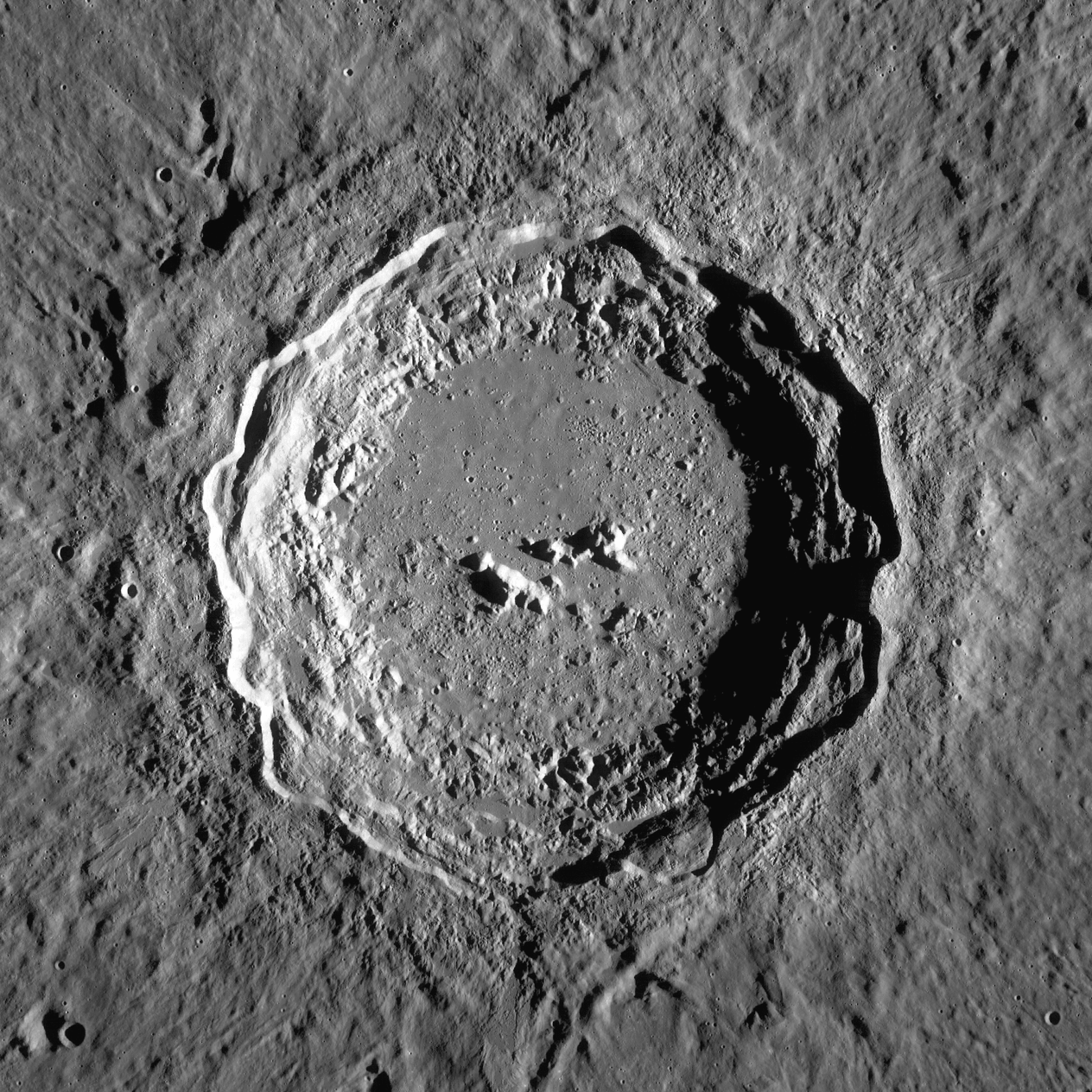}
\caption{A lunar impact crater called \emph{Copernicus} (\textit{the image is avaiable from Wikipedia creative commons}). } 
\label{fig:crater}
\end{figure}

\textbf{Humans, on the other hand, apply reasoning or even meta-reasoning that involves outside information.} For example, let us consider a task of detecting whether a moon crater, such as the one shown in Figure~\ref{fig:crater} on a picture is concave or convex. 
Machine learning techniques can be easily fooled because concave craters with shadows to the left are very similar to convex craters with shadows to the right. If there is a watermark with the time of day the picture was made, humans may notice it spontaneously and infer the direction of the sun and then accurate shadows. If there were no watermarks with a date information in the training, machine learning models would at best ignore them completely or treat as a noise in data. In fact, even if there were some watermarks on images used during training, it is unlikely that the same ML model would learn to classify the image and interpret text information on top of it.

Moreover, humans use a lot of \emph{out-of-the-box} techniques. Let us consider verbal communication ``task” in a natural language. Humans will also incorporate non-verbal communication \cite{19}, anyway. 

We argue that logical reasoning is required to go beyond just mathematical classification or regression represented by machine learning models. Logical conclusions may completely change the process how we interpret what we see, hear, smell etc. Logical reasoning goes beyond fitting new experiences to the known ones. It allows us to reinterpret things and add adapt to new situations dynamically.

\section{Explainable reasoning }

\begin{figure*}[!ht]
\centering
\includegraphics[width=0.8\linewidth]{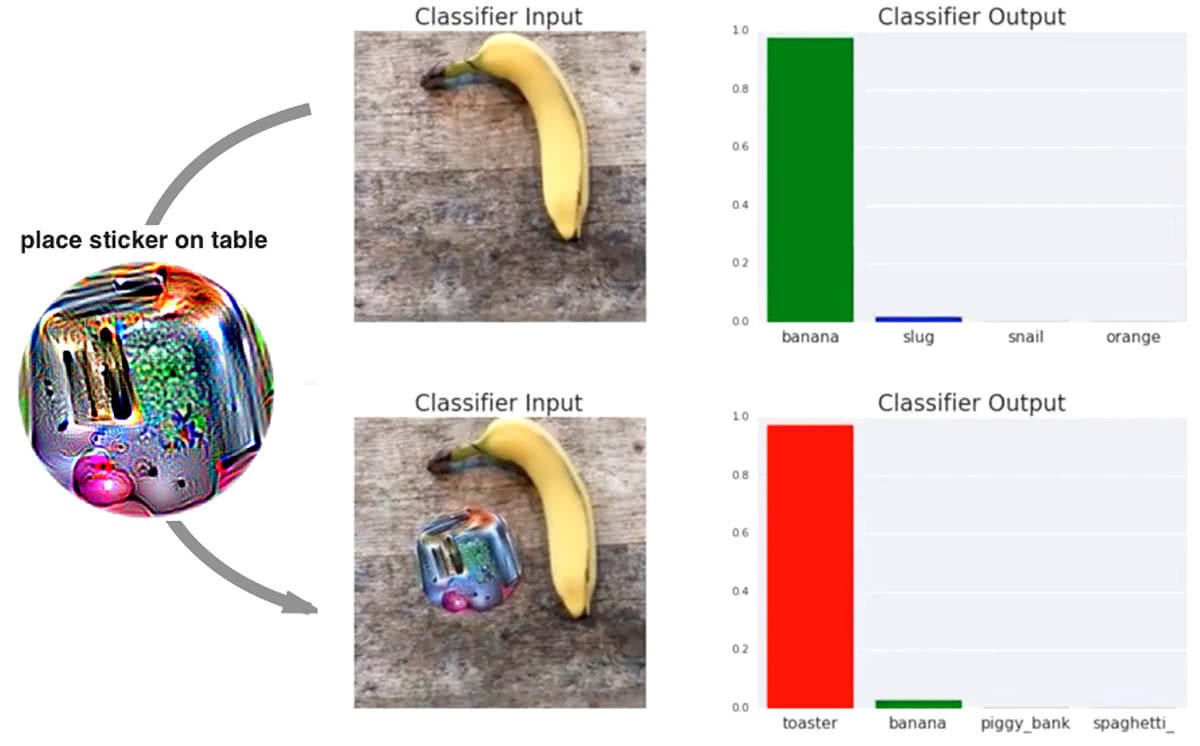}
\caption{The image on the top was classified as a banana by a deep convolutional neural network, whereas the image on the bottom was classified as a toaster. The figure is available in a paper~\cite{brown2017adversarial} published on \emph{arxiv} open repistory. Interested readers are advised to consult this paper for description of the method of how the classifier was fooled.} 
\label{fig:cnn}
\end{figure*}

The proposed definitions of AGI involve some kind of performance metrics at given task and a comparison to human intelligence \cite{3,5}. When attempting to solve a complex task that requires intelligence, \textbf{humans are capable at explaining their thought process} and justify their decisions or actions. In rare occasions when we are not able to, we will say that it was an instinct, impulse or intuition. 

We argue that entities that possess general intelligence should be able to provide a similar form of justification for their decisions/actions. For the purpose of this article, let us call this explainability.

DNNs as well as traditional NNs are machine-learning models that are considered one of the least explainable ones \cite{20}. Even the recent advancements in the so-called eXplainable AI (XAI) \cite{21} give only limited tools such as feature importance, uncertainty, error, examples, visual summaries etc.  The bottom line is that NNs are large numerical function approximators and do not really operate on explainable concepts inherently.

Naturally, there are machine learning models such as \emph{decision trees} \cite{22} that are inherently explainable and interpretable, but they are usually much less effective in solving complex problems, for which deep learning is employed.

\section{Vulnerability to attacks}

A significant part of the overall success of deep learning has been due to Convolutional Neural Networks (CNNs). They are state-of-the-art approaches in many image classification tasks and computer vision~\cite{24}, in general. As discussed in~\cite{25}, CNNs -- in particular -- can be maliciously manipulated by presented specially prepared input. This is referred to as \emph{adversarial attacks}. Naturally, not all examples of fooling deep learning AI have to be malicious~\cite{heaven2019deep}. Imagine a road sign detection AI that is fooled because the view by the camera has been obscured. The are hundreds of example on the Internet, such as the one shown in Figure~\ref{fig:cnn}, that show spectacular misclassifications of images by deep neural networks. These examples are considered funny, because humans do not make such obvious mistakes.

Partial reason for such attacks being possible is that these types DNNs do not operate on particularly meaningful features as input. Such features are discovered by the network within intermediate layers. The input consists in a lot of very simple features – RGB color values for millions of pixels – and the next layers of the network may represent features using mathematical operations on them including weighted aggregation. It is possible to exploit this, for instance, by specifying a completely different image that will result in the same results of internal operations further in the network.

Humans are particularly good at seeing patterns and concepts, not individual pixels. Moreover, humans have all sorts of built-in mechanisms, such as premonition, that prevent them from being fooled so easily but more importantly – we build our judgements in a holistic fashion, also using external data. Human intelligence is not a portfolio of specialized algorithms for particular tasks. \textbf{We use one brain for everything}. Therefore, it is likely that the defence mechanisms go in pair with fully implemented transfer learning. Machine learning models optimized for one problem have limited additional validation possibilities, because there is no ``world'' for them beyond this one problem. They cannot answer, for instance: ``\emph{It looks like cat, but I know this is not a cat, because you have always tried to fool me so far in the history of our interactions''}.

\section{Conclusions}
\label{sec:conclusions}

Deep learning has proven to be able to break barriers in front of AI. One of the most prominent examples is an ancient board game -- \emph{Go} -- which was considered to be one of the ``grand challenges of AI''~\cite{26}. In 2016, \emph{AlphaGo}, a program based on deep neural networks and Monte Carlo Tree Search, defeated Lee Se Dol, who is one of the strongest \emph{Go} players of all time. \emph{AlphaGo} was a major breakthrough in computer science field. This approach has inspired many successful approaches to other games and problems~\cite{28}. 

Through the history of civilization, human intelligence has allowed us to solve numerous problems and overcome many challenges. AGI, to earn its title, also needs to be effective at solving complex problems. It seems like deep learning is a promising way to achieving this goal. 

However, AGI is something more that efficacy at one task. In this short article we have pointed out five characteristics of deep neural network that make us believe that AGI is still far beyond reach of deep learning. 

These characteristics revolve around:
\begin{itemize}
    \item Faster learning, even with limited training examples
    \item Transfer learning and using analogies
    \item Abstract out-of-the-box (logical) reasoning
    \item Deep explanation of the thought process
    \item Defence mechanisms e.g. against malicious behavior
\end{itemize}

It is worth aiming for the improvement in these five areas regardless of AGI. We believe that they all help in creating better and more robust machine learning models.

\section*{Acknowledgements}

This work is a part of the SENSEI project, which was co-funded by Smart Growth Operational Programme 2014-2020, financed by European Regional Development Fund under GameINN project POIR.01.02.00-00-0184/17, operated by National Centre for Research and Development in Poland.

\bibliographystyle{IEEEtran}
\bibliography{main}

\end{document}